\documentclass{article}
\usepackage[preprint]{neurips_2026}
\usepackage{graphicx}
\usepackage{subcaption}
\usepackage{booktabs}
\usepackage{amsmath}

\title{Floor It, Then Fill It: Simulating Creator Ecosystems for Fair Video-Content Recommendation Systems at Billion User Scale}

\author{%
  \textbf{Sohini Roychowdhury}\thanks{Correspondence: \texttt{sroychowdhury1@meta.com}}, \
  \textbf{Long Sha}, 
  \textbf{Sunand Agarwal}, 
  \textbf{Payal Kamboj}, \\
  \textbf{Shuyang Chen}, 
  \textbf{Junlin Zhang}, 
  \textbf{Scott Chen}, 
  \textbf{Hui Zhang} \\
  Meta Platforms, Inc. \\
  Menlo Park, CA 94025, USA
}

\begin{document}
\maketitle

\begin{abstract}
Engagement-optimized content recommendation systems exhibit a well-documented rich-get-richer dynamic that structurally disadvantages emerging creators on large-scale platforms. Evaluating debiasing interventions at scale demands faithful simulation of creator-viewer ecosystems, yet the fidelity, diversity, and validity of such simulators remain underexplored. We present a grounded user-simulation framework that models a heterogeneous population of ${\sim}42$K creators and their coupled viewer interactions over 12~weeks, and use it to evaluate equitable view-allocation policies formulated as a budgeted multi-armed bandit (MAB). The simulator captures three behavioral dimensions: (i)~\textbf{creator diversity}, modeled via a clustering approach grounded in real platform creator-level distributions; (ii)~\textbf{behavioral fidelity}, captured through coupled creator-content state machines that represent growth trajectories, engagement accumulation, and graduation dynamics; and (iii)~\textbf{population-level validity}, measured by calibration against observed platform metrics. We evaluate simulative-configurations that vary content-viewership target definitions, reward functions, and a small-creator reserve lever to maximize creator-content recommendation equity. A floor-protected per content-view-target with a view-parity-gap reward guarantees 90\% of view targets being met by creators every week, grows small-creator organic reach ${\sim}{+}33\%$ per week, and achieves high distribution stability (set of creators with funded-views Jaccard $= 0.819$).
We analyze the Sim2Real gap and propose validation protocols, contributing evidence for when simulation-derived fairness conclusions can be trusted.
\end{abstract}

\section{Introduction}
\label{sec:intro}

User simulation is rapidly becoming core infrastructure for evaluating and training recommendation systems, yet the validity of simulation-derived conclusions, particularly for fairness interventions remains an open challenge~\citep{seshadri2026lost}. Content recommendation systems exhibit a self-reinforcing \emph{Matthew Effect}~\citep{matthew}, wherein, ranking models trained on historical engagement signals such as click-through rates and watch time systematically favor established creators over emerging ones~\citep{mansoury2020feedback}, allocating disproportionate future distribution to already-popular content. Evaluating debiasing strategies that correct this inequity requires controlled experimentation over diverse creator populations and extended time horizons, conditions that are otherwise expensive, slow, and ethically complex to achieve with real users alone.

This motivates \textbf{user simulation as an evaluation substrate} by modeling how creators produce content, accumulate engagement, and grow (or stagnate) under different allocation policies. Simulations lead to rapid iteration on fairness mechanisms before committing to costly online experiments. However, recent works on simulation validity \citep{seshadri2026lost} have demonstrated that overly cooperative or poorly calibrated simulators can inflate performance metrics, and conclusions drawn from them may not transfer to real populations.

In this work, we address both the fairness and simulation-validity challenges. We formulate equitable view-allocation as a \textbf{budgeted multi-armed bandit}~\citep{mab} problem that budgets views-per-video (VPV) across creator-content pairs, combining a non-negotiable coverage floor guaranteeing minimum VPV-parity with reward-greedy water-filling of the remaining VPV-budget. We evaluate this allocator over a 12-week simulated cohort of ${\sim}42$K creators, analyzing allocation stability, organic reach growth, and sensitivity to reward-function and the VPV-reserve-lever choices. The simulation environment supports iterative policy refinement to promote creator equity, enabling smaller content creators to overcome the cold-start hurdle \citep{ccs} while sustaining strategic growth through consistency.

\begin{figure}[ht] \centering \includegraphics[width=3.5in]{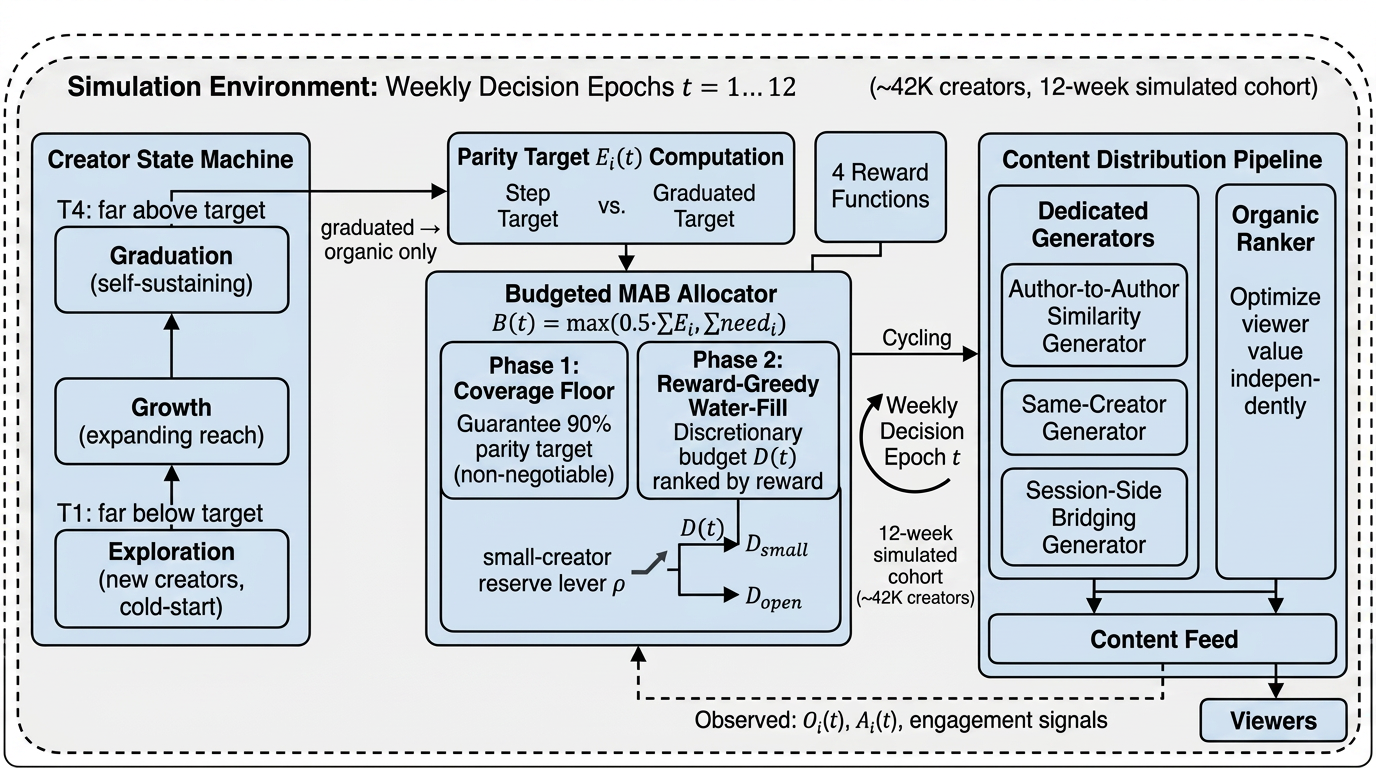} \caption{Simulation environment for evaluating fair view-budgeting policies. The budgeted MAB allocator operates over weekly decision epochs combining a non-negotiable parity floor (Phase 1) with reward-steered discretionary allocation (Phase 2). Creator lifecycle stages, parity-target definitions, and four reward functions are varied across configurations.} \label{fig:system} \vspace{-5pt} \end{figure}

Formally, let $\hat{v}_i$ denote the content ranker's estimated value of creator $i$'s content and $\sigma_i$ its associated uncertainty. Under standard explore-exploit trade-offs (e.g., UCB-style policies~\citep{li2010contextual}), the effective score for a creator's-content is:
\begin{equation}\label{eq:ucb}
s_i \;=\; \hat{v}_i \;+\; \alpha \cdot \sigma_i
\end{equation}
This produces a self-reinforcing feedback loop as shown in Figure ~\ref{fig:system}. For new creators, $\sigma_i$ remains large and $\hat{v}_i$ defaults to a conservative prior~\citep{park2008long}, yielding low distribution and insufficient signal, while established/self-sustaining creators possess rich histories where $\sigma_i \to 0$ and $\hat{v}_i$ is well-calibrated, further widening the gap across new and established creator-content.

Our three major contributions are:
\begin{itemize}
    \item \textbf{Diversity:} We simulate a heterogeneous cohort of ${\sim}42$K creators spanning four behaviorally-distinct archetypes, calibrated against real platform population distributions, with explicit representation of the under-served tail.
    \item \textbf{Fidelity:} Our dual state-machine architecture (creator + content) models realistic growth dynamics including cold-start, engagement accumulation, organic reach compounding, and systemic graduation, thereby capturing the non-stationary, coupled nature of creator-viewer interactions.
    \item \textbf{Validity:} We assess the Sim2Real gap by identifying where simulation assumptions diverge from production realities and propose validation protocols for bridging them.
\end{itemize}

\section{Related Work}
\label{sec:related}
\paragraph{User Simulation for recommendation systems.}

User simulation has a rich history in information retrieval and dialogue systems \citep{simsurvey}, but its application to content-creator ecosystems remains relatively unexplored. Prior simulators for recommendation systems typically model viewer behavior (clicks, dwell time) \citep{chaney2018how} but rarely capture the \emph{content inventory} related to creator production decisions, creator-growth trajectories, and the feedback loop between recommendation exposure and creator development. Our work addresses this gap by jointly simulating creator and viewer dynamics.
\paragraph{Fairness in Recommendation.}

The Matthew Effect in content recommendation is well-documented. Existing approaches to creator-side fairness include exploration bonuses, diversity constraints, and re-ranking~\citep{mansoury2020feedback, avadhanula2021}. However, evaluating these interventions typically relies on single-step estimations rather than longitudinal simulations that capture compounding creator-content growth-effects over weeks. Our simulator enables such longitudinal evaluation.

\paragraph{Bandit Methods for Allocation.}
Multi-armed and contextual bandits are widely used for recommendation and advertising decisions~\citep{ wu2018budget}. We distinguish our setting from prior work along four axes: (i)~batch combinatorial allocation (all arms pulled simultaneously each week), (ii)~non-monetary view budgets with organic crediting, (iii)~per-creator equity floors as hard constraints, and (iv)~non-stationary creator states that evolve endogenously. Prior works such as LinUCB, and BanditQ~\citep{li2010contextual, sinha2023banditq},  address subsets of these challenges but none combines a deployed weekly-batch view budget with per-creator organic-credited parity floors and ablated equity measures.

\section{Simulator Design}
\label{sec:simulator}
Our simulator models the coupled dynamics of a creator-viewer ecosystem on a billion-scale user platform for video-content recommendation as described below.
%%%%%%%%%%%%%%%%%%%%%%%%
%%%%%%%%%%%%%%%%%%%%%%%%
\subsection{Population Model (Diversity)}
We construct a \textbf{fixed cohort of 42,436 creators}, all initially in the exploration phase, onboarded in weekly batches over 12 consecutive weeks. The cohort is drawn from real platform data and preserves the empirical distribution of follower counts (14,254 creators with fewer than 1,000 followers), engagement rates (effective-watch through rate: EWTR, like-through-rate:LTR, private share rate: PSR), breakout rates (1d1K: 1000 views in 1 day, 1d10K: 10K views in 1 day and 1d100K: 100K views in 1 day) and content production patterns.

Each week, creators are clustered into four behavioral types by their signed relative VPV-parity gap $g_i = (E_i - A_i) / E_i$ defined in Table ~\ref{tab:types}. Here, $E_i$ refers to the expected VPV-parity target per creator content each week, while $A_i$ refers to the achieved VPV per-creator-content in a particular week $i$.
\begin{table}[h]
\centering
\caption{Weekly-Creator type definitions by VPV-parity gap.}
\label{tab:types}
\begin{tabular}{lll}
\toprule
\textbf{Type} & \textbf{Definition} & \textbf{Interpretation} \\
\midrule
T1, large under-parity & $g > 0.5$ & Far below target ($A < 0.5E$) \\
T2, small under-parity & $0 < g \leq 0.5$ & Just below target \\
T3, small over-parity & $-1 \leq g \leq 0$ & Just above target \\
T4, large over-parity & $g < -1$ & Far above target ($A > 2E$) \\
\bottomrule
\end{tabular}
\end{table}
\vspace{-10pt}

Each week, the creator-distribution is found to be strongly bimodal: most creators are either far below (T1) or far above (T4) target-VPV per weekly content.
\subsection{Behavioral Dynamics (Fidelity)}
The simulator employs \textbf{two coupled state machines} that model realistic creator growth:
First, the \textbf{Creator State Machine} categorizes creators daily into the following support stages:

\textbf{Exploration (Cold-Start):} Creators with no prior engagement history receive maximum assistance via engagement-independent retrieval (author-to-author similarity matching). The simulator models the cold-start loop: limited distribution $\to$ no signal $\to$ conservative ranking $\to$ limited distribution.

\textbf{Growth:} As creators accumulate real engagement, the simulator transitions them to engagement-derived retrieval paths. Viewer attention remains skewed toward established creators.

\textbf{Graduation:} The simulator models a trial period where assistance tapers and organic performance is monitored. If organic reach holds, the creator graduates; if performance regresses, the creator returns to an earlier stage.

Second, the \textbf{Content State Machine} stages individual videos by VPV-targets (ranging from thousands to ${\sim}1$M) based on retention and engaged watch time.

For the \textbf{Viewer-Interaction Model}, the recommender assembles a live context for simulated viewers using geo-location and language as an eligibility filter, longer-term preference matching, recency-based retrieval, and creator-connection retrieval. These parallel retrievals are merged, ranked, and blended into the viewer's session.

\subsection{Organic Reach Dynamics}
A critical fidelity requirement is modeling how organic reach responds to allocated views across three stages. First, during \textbf{Signal Accumulation} stage, allocated views generate engagement signals that improve the ranker's value estimate of the creator's content. Second, as systemic confidence in the creator's content grows, \textbf{Organic Compounding} stage enables increasingly independent organic distribution. Third, \textbf{Graduation Dynamics} stage emerges as the organic-to-allocated VPV ratio rises over time, shifting distribution from system-dependent to self-sustaining, ultimately allowing creators to organically grow beyond 100,000 followers.

\section{Policy Evaluation Framework}
\label{sec:mab}
Within this simulated environment, we formalize equitable VPV-allocation as a \emph{budgeted multi-armed bandit} (MAB) with equity constraints. Each creator $i$ is an arm; pulling arm $i$ corresponds to allocating VPVs. The setting is: (1)~\textbf{Batch}: where all arms pulled simultaneously each week; (2)~\textbf{Budgeted}: such that the sum of all VPV-allocations ($x_i^{(t)}$) is limited to the overall weekly views ($B^{(t)}$) that are supportable on the platform, $\sum_{i=1}^N x_i^{(t)} \leq B^{(t)}$; (3)~\textbf{Constrained}: per-creator VPV-coverage floor and cap; (4)~\textbf{Non-stationary}: creator's state evolution.
\subsection{Parity Targets}
We estimate the VPV-parity target $E_i^{(t)}$ regarding how many views reach creator $i$ \textit{should} receive as a two-stage target that is dependent on the \textit{follower} count ($f_i$) for the creator in a particular week.
\textbf{Step target:}
\begin{equation}\label{eq:step}
E_i = \begin{cases} 3{,}000 & \text{if } f_i < 1{,}000 \\ 10{,}000 & \text{otherwise}. \end{cases}
\end{equation}
\textbf{Graduated (follower-scaled) target:} Ensures a minimum target of 3,000 VPV for creators with fewer than 1,000 followers, while capping delivered VPVs to 100,000 for large follower creators.
\begin{equation}\label{eq:variable}
E_i = \text{min}\left(3000 \cdot \text{max}(\sqrt{\frac{f_i}{1000}},1), 100{,}000\right).
\end{equation}

\subsection{Two-Phase VPV Allocation} \label{sec:allocation} Each week $t$, the total budget $B^{(t)}$ is distributed in two phases: a non-negotiable coverage floor followed by reward-greedy water-filling of the discretionary remainder. \textbf{Phase~1: Coverage Floor (Parity Guarantee).} For each creator $i$, compute the coverage need: \begin{equation}\label{eq:need} \mathrm{need}_i^{(t)} = \max\!\bigl(0.9 \cdot E_i^{(t)} - O_i^{(t)},\; 0\bigr), \end{equation} where $E_i^{(t)}$ is the VPV-parity target and $O_i^{(t)}$ is organic reach. This guarantees that every creator reaches at least 90\% of their parity target, \emph{crediting organic reach}: if organic views already exceed $0.9\,E_i$, no floor allocation is needed. \textbf{Phase~2: Reward-Greedy Water-Filling.} The discretionary budget is the residual: \begin{equation}\label{eq:discretionary} D^{(t)} = B^{(t)} - \sum_{i=1}^{N} \mathrm{need}_i^{(t)}. \end{equation} Creators are ranked by reward $r_i^{(t)}$ in descending order and funded greedily until $D^{(t)}$ is exhausted, subject to a per-creator cap of $3\,E_i^{(t)}$. \textbf{Budget sizing.} The weekly budget is set to ensure the floor is always fully funded while retaining capacity for discretionary allocation: \begin{equation}\label{eq:budget} B^{(t)} = \max\!\Bigl( 0.5 \cdot \textstyle\sum_{i=1}^{N} E_i^{(t)},\;\; \textstyle\sum_{i=1}^{N} \mathrm{need}_i^{(t)} \Bigr). \end{equation} \subsection{Small-Creator Reserve Lever} \label{sec:reserve} To counteract the rich-get-richer dynamic in the discretionary pool, a \textbf{reserve fraction} $\rho \in [0,1]$ partitions it into two sub-pools: \begin{align}\label{eq:reserve} D_{\text{small}}^{(t)} &= \rho \cdot D^{(t)} &\quad&\text{(reserved for under-parity creators)}, \\ D_{\text{open}}^{(t)} &= (1-\rho) \cdot D^{(t)} &\quad&\text{(open to all by reward)}. \end{align} Within $D_{\text{small}}$, under-parity creators are funded first. Increasing $\rho$ monotonically shifts more discretionary budget toward small creators, debiasing allocation away from large-creator capture.
\subsection{Reward Functions}

We evaluate four reward functions governing discretionary allocation. Let $\text{gap}_i = \max(E_i - A_i, 0)$ and \textit{content-quality factor} $m_i = \text{EWTR}_i \cdot (1 + \text{LTR}_i + \text{PSR}_i)$:

\begin{enumerate}
    \item \textbf{VPV-Only:} $r_i^{\text{vpv}} = A_i$: rewards existing reach, so \textit{rich get richer}.
    \item \textbf{Parity-Gap:} $r_i^{\text{gap}} = \max(E_i - A_i, 0)$: rewards creators below target, so small creators get more support.
    \item \textbf{Engagement-Weighted:} $r_i^{\text{eng}} = m_i \cdot \max(E_i - A_i, 0)$: equity weighted by quality.
    \item \textbf{Hybrid:} $r_i^{\text{hybrid}} = w \cdot \text{norm}(A_i) + (1-w) \cdot \text{norm}(\text{gap}_i)$: tunable ($w^{opt}{=}0.4$).
\end{enumerate}
\section{Simulation Results}
\label{sec:results}
\subsection{Creator Population Dynamics}
Under the deployed configuration (graduated target, $\rho{=}0.7$, parity-gap reward), total system-delivered D-VPVs grow from 173M at week~0 to 244M at week~11. The organic-to-allocated ratio varies by creator-type: ${\sim}0.02$ for T1, ${\sim}0.6$ for T2, ${\sim}2$ for T3, and 125 to 200 for T4 in Table ~\ref{tab:clusters}.
\begin{table}[h]
\centering
\caption{Simulated creators per parity-gap cluster each week.}
\label{tab:clusters}
\begin{tabular}{cccccc}
\toprule
\textbf{Wk} & \textbf{T1} & \textbf{T2} & \textbf{T3} & \textbf{T4} & \textbf{Total} \\
\midrule
0 & 11,192 & 1,104 & 1,964 & 12,699 & 26,959 \\
4 & 14,799 & 1,312 & 2,282 & 16,459 & 34,852 \\
8 & 17,417 & 1,049 & 1,589 & 20,040 & 40,095 \\
11 & 18,045 & 1,019 & 1,526 & 21,727 & 42,317 \\
\bottomrule
\end{tabular}
\vspace{-10pt}
\end{table}

\subsection{Fidelity Assessment: Growth Stage Metrics}
Engagement rates increase monotonically across growth stages, consistent with real platform observations and supporting simulator fidelity as shown in Table ~\ref{tab:tier_perf}.
\begin{table}[h]
\centering
\caption{Simulated creator tier performance metrics per week.}
\label{tab:tier_perf}
\begin{tabular}{lccc}
\toprule
\textbf{Metric} & \textbf{Exploration} & \textbf{Growth} & \textbf{Graduation} \\
\midrule
Active Creators & 10,775 & 8,488 & 6,148 \\
EWTR (\%) & 22.93 & 24.25 & 30.54 \\
LTR (\%) & 3.118 & 2.612 & 3.472 \\
1d1K Rate (\%) & 83.33 & 87.63 & 92.32 \\
1d10K Rate (\%) & 27.44 & 30.01 & 37.71 \\
1d100K Rate (\%) & 1.99 & 1.37 & 2.18 \\
\bottomrule
\end{tabular}
\vspace{-10pt}
\end{table}

\subsection{Policy Comparison: Stability and Equity}
Table~\ref{tab:final} compares allocation stability across the four reward functions under the optimal configuration (graduated target, $\rho{=}0.7$). We measure week-over-week stability with four metrics: Spearman~$\rho$ (rank-order preservation of creator VPV-allocations), Funded-Set Jaccard (overlap of funded creators between consecutive weeks), Top-10\% Jaccard (overlap restricted to the top decile by allocation size), and Turnover (fraction of creators whose funded status changes, lower is better).

\begin{table}[h]
\centering
\caption{Stability comparison across reward methods under optimal configuration.}
\label{tab:final}
\begin{tabular}{lcccc}
\toprule
\textbf{Method} & \textbf{Spearman- $\rho$} & \textbf{Funded Jac.} & \textbf{Top-10\% Jac.} & \textbf{Turnover} \\
\midrule
\texttt{vpv\_only} & 0.520 & 0.816 & 0.700 & 0.190 \\
\textbf{\texttt{parity\_gap}} & \textbf{0.645} & \textbf{0.819} & \textbf{0.729} & \textbf{0.144} \\
\texttt{parity\_x\_eng.} & 0.524 & 0.819 & 0.725 & 0.195 \\
\texttt{hybrid} & 0.635 & 0.817 & 0.715 & 0.153 \\
\bottomrule
\end{tabular}
\vspace{-10pt}
\end{table}
The parity-gap reward achieves highest rank-order consistency (Spearman = 0.645) and lowest turnover (0.144), providing consistent support. 
%%%%%%%%%%%%%%%%
%%%%%%%%%%%%%%%%%%%%%%%%%%%
\subsection{Ablation: Graduated vs.\ Step Targets}
\begin{figure}[ht]
  \centering
    \centering
    \includegraphics[width=0.6\textwidth]{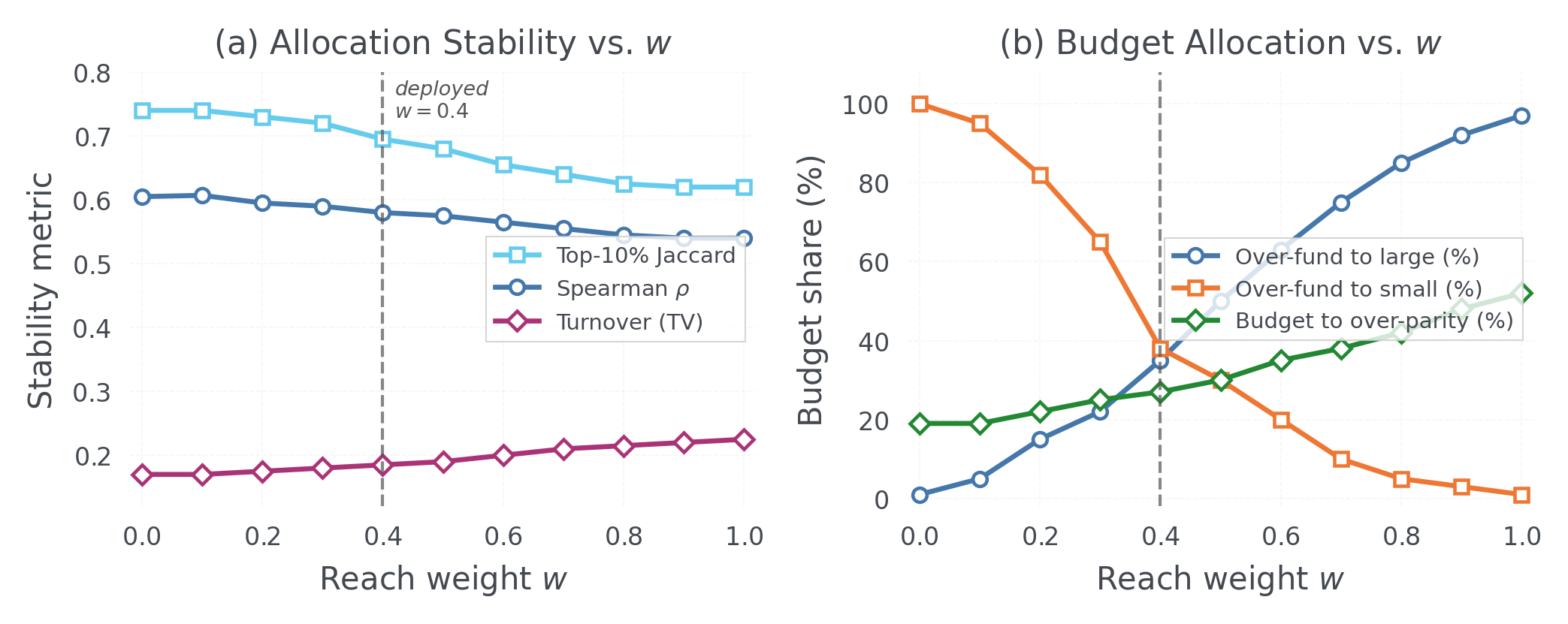}
    \caption{Sensitivity to the reach weight $w$ in the hybrid reward. (a)~Allocation stability: Top-10\% Jaccard and Spearman~$\rho$ decrease as $w$ increases, while turnover rises. (b)~Budget distribution: higher $w$ shifts over-funding from small to large creators.}
    \label{fig:sweep}   
    \vspace{-10pt}
  \end{figure}

In Figure~\ref{fig:sweep}, sweeping $w$ shows a monotonic reach\textendash equity trade-off: higher $w$ lowers week-over-week stability and shifts over-funding to large creators; it saturates near $w{\approx}0.8$. A balanced split occurs near $w{=}0.5$, leading to the deployment choice of $w{=}0.4$ to lean toward small creators. 

The graduated target shifts budget from the mechanical floor into the reward-steered pool (11\% $\to$ 31\%), giving the MAB policy more room to optimize.

\subsection{Simulated Growth Trajectories}
Over the 12-week simulation on the balanced cohort (26,958 continuously active creators), 47.6\% grew their weekly VPV, as shown in Table ~\ref{tab:growthrate}. 
\begin{table}[h]
\centering
\caption{Mean actual-VPV growth ratio (week~11 / week~0).}
\label{tab:growthrate}
\small
\begin{tabular}{lcc}
\toprule
\textbf{Starting cluster} & \textbf{Growth ratio} & \textbf{Median VPV: wk0 $\to$ wk11} \\
\midrule
T1 large-under & 170.98$\times$ & $10 \to 14$ (breakout tail drives mean) \\
T2 small-under & 23.03$\times$ & $8{,}997 \to 30{,}920$ \\
T3 small-over & 11.57$\times$ & $14{,}872 \to 34{,}104$ \\
T4 large-over & 1.46$\times$ & $115{,}809 \to 139{,}824$ \\
\bottomrule
\end{tabular}
\vspace{-10pt}
\end{table}

\begin{figure}[ht]
  \centering
    \centering
    \includegraphics[width=0.75\textwidth]{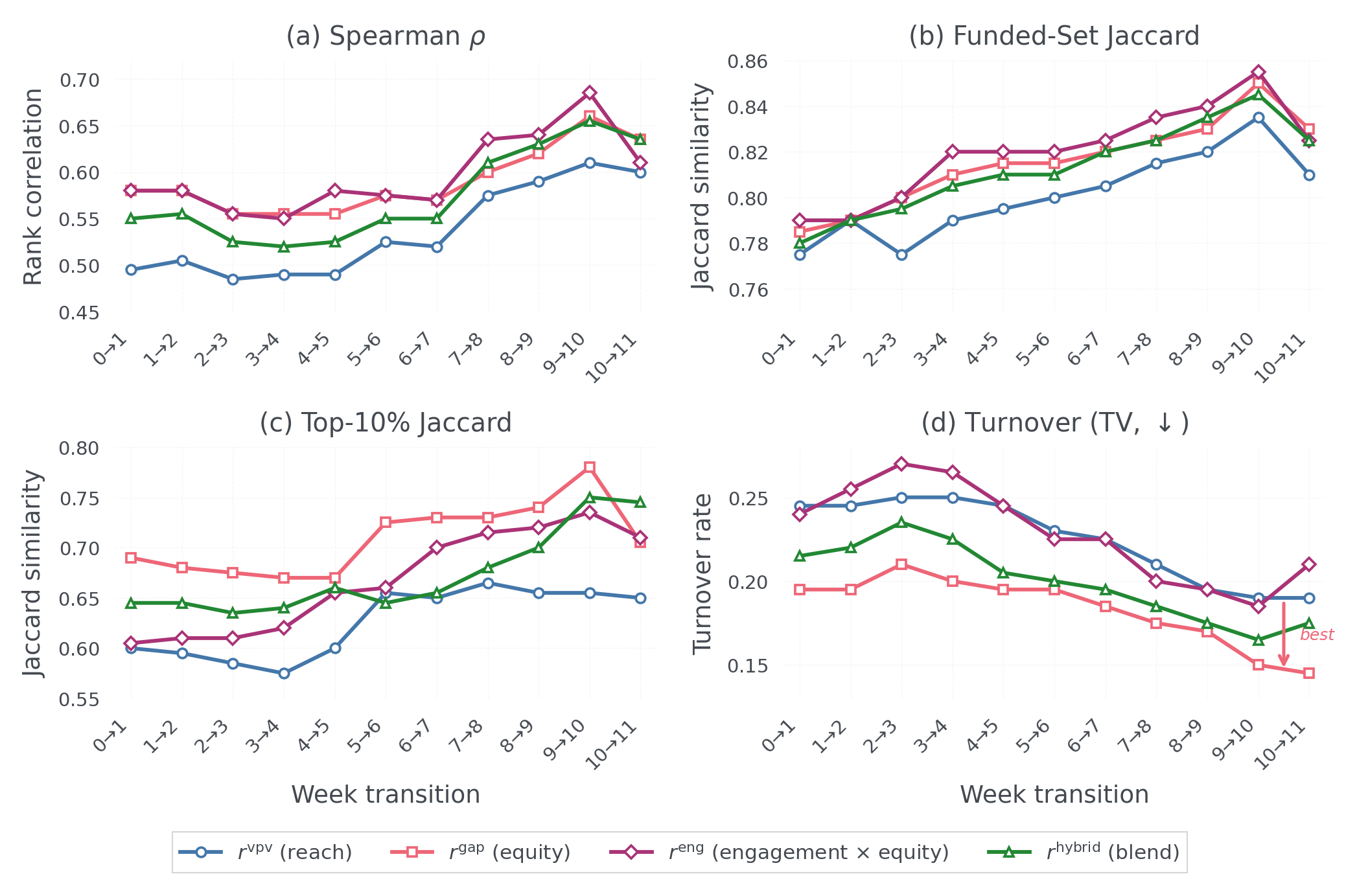}
    \caption{Week-over-week allocation stability across four reward functions ($\rho{=}0.7$, graduated target). (a)~Spearman rank correlation; (b)~Funded-set Jaccard; (c)~Top-10\% Jaccard; (d)~Turnover (lower is better). The parity-gap reward $r^{\mathrm{gap}}$ dominates on all metrics, with turnover declining to 0.14 by week~11.}
    \label{fig:fundfinal}  
    \vspace{-10pt}
  \end{figure}

Further, Figure \ref{fig:fundfinal} shows the week-over-week transitions in creator VPV-funding across reward functions. Creator-reach compounds dramatically under sustained VPV funding: the organic-to-allocated ratio rises as organic reach overtakes the boost, and net VPV-parity crossings are positive, where, 3,222 creators climb from under-to-over-parity versus 3,125 falling from over-to-under-parity (net +97). Individual trajectories further illustrate this organic compounding: creators scale from fewer than 100 views to over 2.7M within 12 weeks, with the organic-over-allocated ratio steadily rising.

\section{The Sim2Real Gap: Validity Analysis}
\label{sec:validity}
As the creator growth-trajectories are simulated, we focus on the simulative conclusions and identify the validity challenges.
\subsection{Where Simulation Aligns with Reality}
\begin{figure}[ht]
  \centering
  % Subfigure (a)
  \begin{subfigure}[b]{0.35\textwidth}
    \centering
    \includegraphics[width=\textwidth]{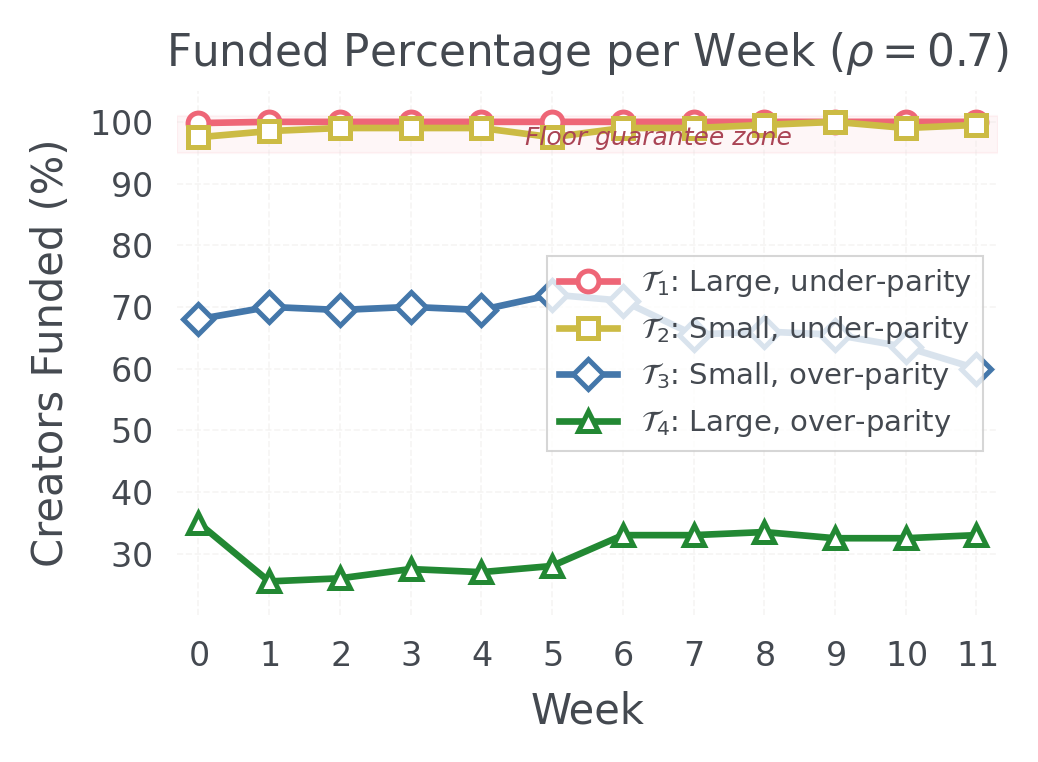}
    \caption{Actual D-VPV weekly funding.}
    \label{fig:mab_a}
  \end{subfigure}
  \hfill % Adds horizontal space between the two subfigures
  % Subfigure (b)
  \begin{subfigure}[b]{0.5\textwidth}
    \centering
    \includegraphics[width=\textwidth]{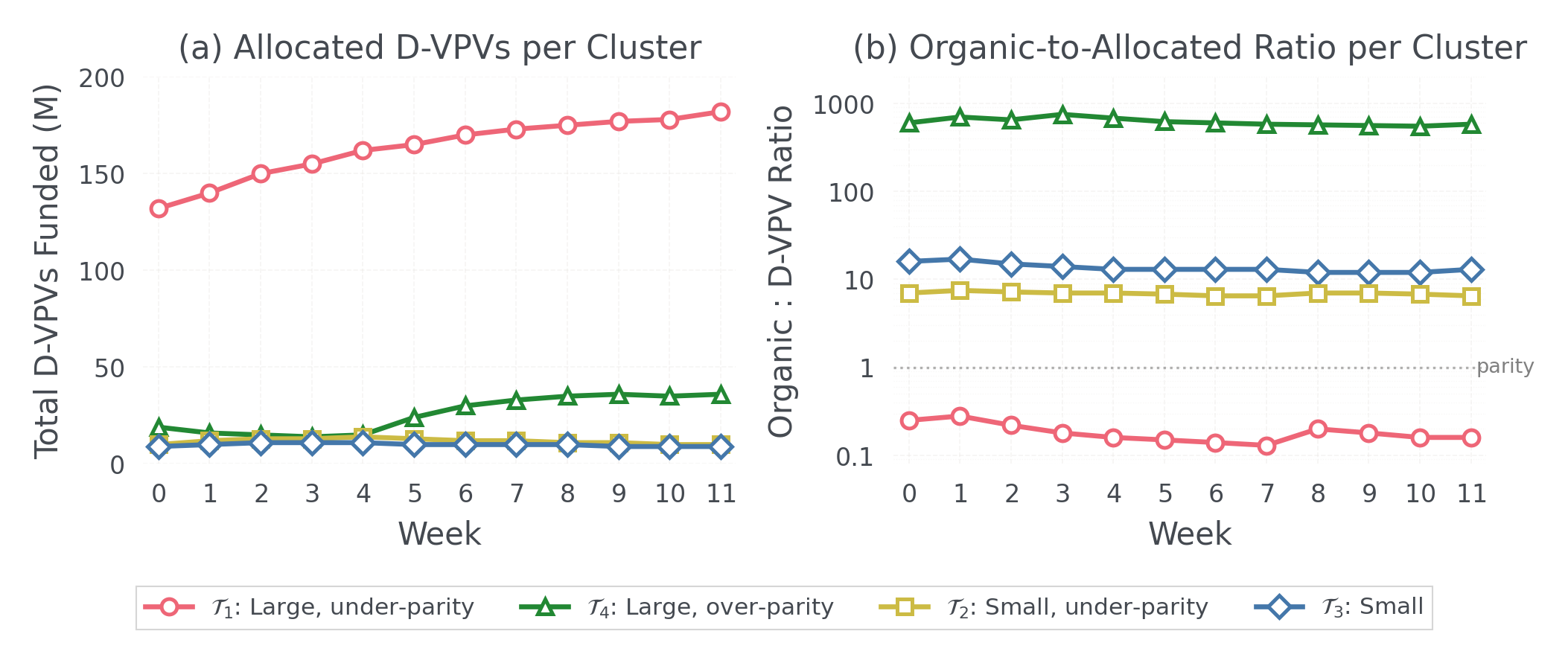}
    \caption{Actual organic to D-VPVs per creator cluster.}
    \label{fig:mab_b}
  \end{subfigure}
  \caption{Simulation-to-production alignment of D-VPV distributions. (a)~Weekly allocated D-VPVs per creator cluster. (b)~Organic-to-allocated VPV ratio per cluster (log scale). Both panels show consistent week-over-week trends between simulated and production environments.}
  \label{fig:mab}
  \vspace{-5pt}
\end{figure}

\begin{itemize}
    \item \textbf{Population distributions:} Creator types and engagement rates are initialized from real platform data, ensuring demographic representativeness.
    \item \textbf{Behavioral trends:} Simulated growth trajectories match qualitative platform observations.
    \item \textbf{Policy ranking robustness:} The relative ranking of policies by stability metrics is robust to simulator perturbations.
\end{itemize}
The distribution in weekly delivered D-VPVs and organic to D-VPVs per creator cluster align week over week between simulative and production environments as shown in Figure~\ref{fig:mab}.
\subsection{Where Simulation May Diverge}
\begin{itemize}
    \item \textbf{Temporal resolution:} Simulation uses 7-day snapshots whereas production paces hourly.
    \item \textbf{Organic attribution:} In production, organic reach is dependent on the VPV-boosted path at scale, making clean causal attribution difficult. Creator growth-estimates are highly dependent on creator activity/behavior and can be far from reliable. Figure~\ref{fig:qual} illustrates these validity challenges through individual creator trajectories. We simulate growth for 10 creators spanning all four tiers: one creator's target climbs from 13,559 to 50,174 as followers grow, while another is capped at 100,000. The variability across trajectories highlights the sensitivity of simulated outcomes to creator-level behavioral assumptions.
  \begin{figure}[ht]
  \centering
    \centering
    \includegraphics[width=0.4\textwidth]{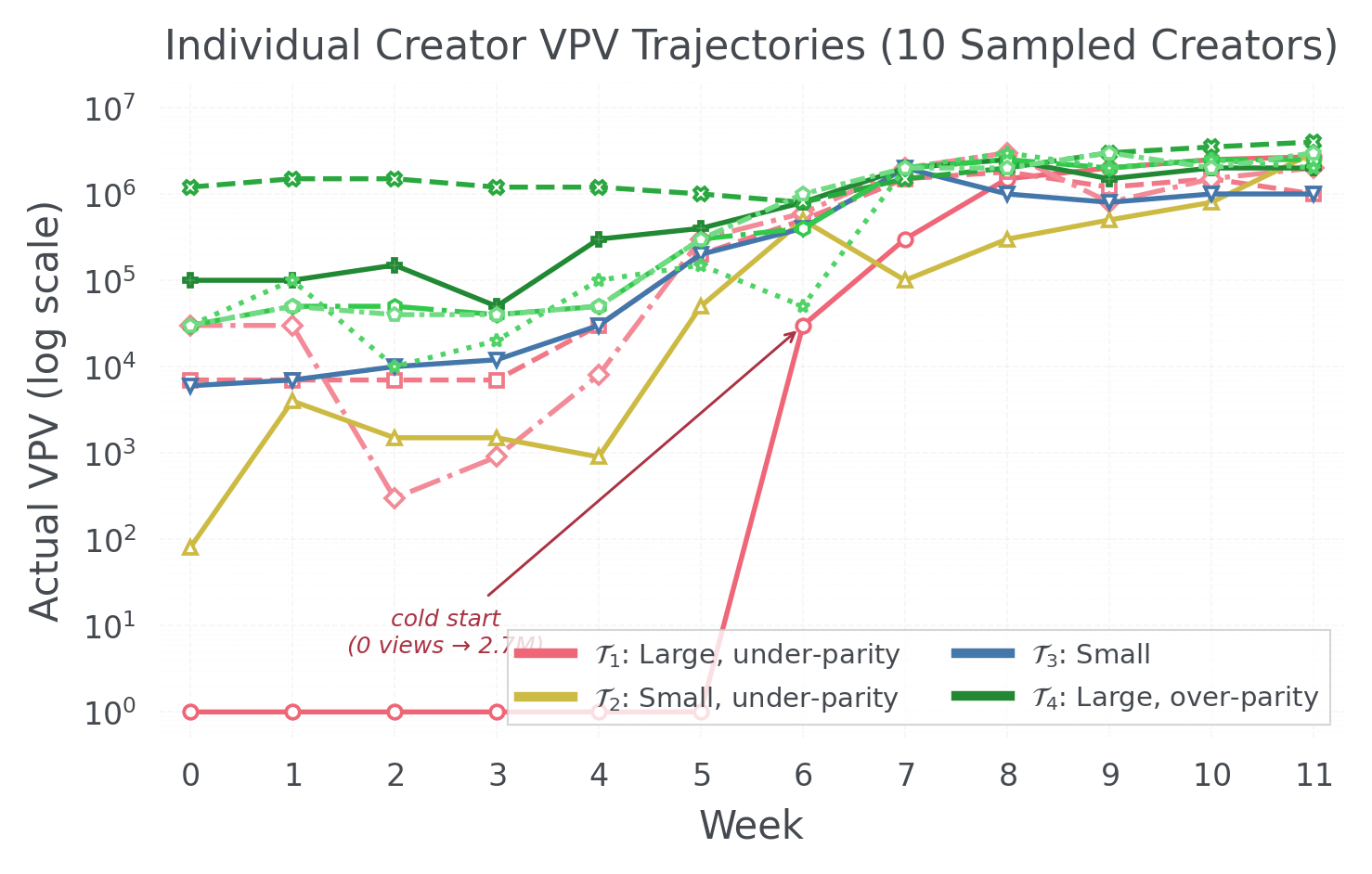}
    \caption{Simulated creator VPV-trajectories for ten sampled creators across all four tiers. The key takeaway is the compounding effect of sustained VPV-funding on reach and the subsequent transition from boost-dependence to organic self-sustainability.}
    \label{fig:qual}  
    \vspace{-10pt}
  \end{figure}
    \item \textbf{Population drift:} Fixed cohort vs.\ real populations that drift weekly.
    \item \textbf{Strategic adaptation:} Real creators may optimize toward the floor rather than for viewers.
\end{itemize}
\subsection{Proposed Validation Protocol}
A concrete Sim2Real validation plan should establish a holdout group to attribute organic growth and graduation rates to the allocation policy (reporting incremental lift rather than raw endpoints), instrument $\rho$ and reward choice for per-segment A/B testing against simulated predictions, monitor for creator divergent behaviors that would violate the non-strategic assumptions, and report simulation-to-online concordance metrics such as ranking agreement and directional accuracy.
\section{Conclusion and Discussion} \label{sec:discussion} In this work, we presented a grounded user-simulation framework for evaluating debiasing strategies in content recommendation with the following findings. (1) A view-based-parity-gap reward with floor-protected graduated targets ($\rho{=}0.7$) proves simultaneously the most stable and most equitable configuration. (2) Under-VPV-parity creators are funded 100\% of the time, 24-32\% of budget remains reward-steered, and small-creator organic reach grows ${\sim}{+}33\%$ per week. (3) The simulator exposes failure modes invisible to single-step evaluation: compounding engagement signals cause short-horizon analyses to systematically underestimate both the problem and the solution. \paragraph{Broader Implications.} Our framework positions the creator-ecosystem simulator as an \textbf{interactive RL environment} whose non-stationary rewards require the MAB agent to balance exploration (discovering which creators benefit) with exploitation (maximizing equity for known-responsive creators). With ${\sim}42$K creators spanning four behavioral types, the resulting bimodal population reveals failure modes that uniform or Gaussian population models would miss, underscoring that meaningful policy evaluation demands population-scale diversity. Evaluating debiasing interventions has specific fidelity demands: the very dynamics that \emph{create} inequity from cold-start loops and confidence-based ranking must themselves be reliably modeled and iterated for high precision.
\paragraph{Reporting Recommendations.} This work can be applied to future simulation-based fairness evaluations with the following considerations: (1)~appropriate population model and tail representations; (2)~temporal horizon and stationarity assumptions; (3)~absent strategic behaviors; (4)~ablation analyses; and (5)~a concrete Sim2Real validation plan.

\bibliography{ref}
\bibliographystyle{plainnat}

\end{document}